\RequirePackage{fix-cm}
\documentclass[smallextended,natbib]{svjour3}       
\smartqed  
\usepackage{parskip}
\usepackage{graphicx}

\usepackage{caption} 
\usepackage{subcaption}
\usepackage{amsmath}
\usepackage{amsfonts}
\usepackage{amssymb} 
\usepackage{enumerate} 
\usepackage{bbm} 
\usepackage{url}

\DeclareMathOperator*{\argmax}{arg\,max}
\begin{document}

\title{Extending counterfactual accounts of intent to include oblique intent
}


\author{Hal Ashton       
}


\institute{H. Ashton \at
              University College London \\
              \email{ucabha5@ucl.ac.uk}
}


\maketitle

\begin{abstract}
One approach to defining Intention is to use the counterfactual tools developed to define Causality. Direct Intention is considered the highest level of intent in the common law, and is a sufficient component for the most serious crimes to be committed. Loosely defined it is the commission of actions to bring about a desired or targeted outcome. Direct Intention is not always necessary for the most serious category of crimes because society has also found it necessary to develop a theory of intention around side-effects, known as oblique intent or indirect intent. This is to prevent moral harms from going unpunished which were not the aim of the actor, but were natural consequences nevertheless. This paper uses a canonical example of a plane owner, planting a bomb on their own plane in order to collect insurance, to illustrate how two accounts of counterfactual intent do not conclude that murder of the plane's passengers and crew were directly intended. We extend both frameworks to include a definition of oblique intent developed in \cite{Ashton2021DefinitionsAlgorithms}.
\end{abstract}

\section{Introduction}\label{intro}
This paper will take a canonical example of oblique intent, found in \cite{TheLawCommission2015ReformReport} and apply it successively to the definitions of intent found in  \citet{Halpern2018TowardsResponsibility} and \cite{Kleiman-Weiner2015InferenceMaking}. It will suggest a definition of oblique intent in each of their model paradigms.

\section{Oblique or Indirect intent in common law}

Oblique or indirect intent refers to the intentional state of side effects of directly intended actions. Its existence can be justified by the following example found in \cite{TheLawCommission2015ReformReport}: \begin{quote}
    D places a bomb on an aircraft, intending to collect on the insurance. D does not act with the purpose of causing the death of the passengers, but knows that their death is virtually certain if the bomb explodes.
\end{quote} 

Oblique intent can be sufficient to prove murder in the UK which otherwise requires direct intent. This is made clear in the current accepted direction to be made to Juries in England and Wales with respect to Oblique intent, originally formulated in R v Woollin\footnote{\label{woollin}R v Woollin [1999] 1 A.C. 82.}, is as follows:
\begin{quote} The jury should be directed that they are not entitled to infer the necessary intention, unless they feel sure that death or serious bodily harm was a virtual certainty (barring some unforeseen intervention) as a result of the defendant's actions and that the defendant appreciated that such was the case.
\end{quote}

The definition of Oblique intent has proved problematical in the law, with much time spent deliberating about whether its presence is sufficient for the most serious of crimes such as murder \citep{Williams1987ObliqueIntention}. The direction above is the product of an enormous amount of debate. Clause 14 of \cite{Commission2015AppendixBill} clarifies it further:

\begin{quote}
    A person acts intentionally with respect to a result if... although it is not his purpose to cause it, he knows that it would occur in the ordinary 
course of events if he were to succeed in his purpose of causing some other result.
\end{quote}

The important modification here, is that one frame of reference, in which to consider oblique intent, is the world where an intended result is realised. This is to make the definition consistent with that of direct intent which does not consider long shot results less intentional than almost sure ones.

In the USA, the Model Penal Code \citep{TheAmericanLawInsitute2017GeneralCulpability} defines four levels of culpability: Purpose, Knowledge, Recklessness and Negligence. It has been adopted at least in part by many states, and is typically cited as the authority on intention\footnote{Federally prosecuted crimes do not have analogous guidance to follow, and follow case law as they are forced to in the UK}. The definition of acting purposely is analogous to the guidance on oblique intent:

\begin{quote}
    A person acts knowingly with respect to a material element of an offense when: ...if the element involves a result of his conduct, he is aware that it is practically certain that his conduct will cause such a result.
\end{quote}

In the next section we will give a formal setup for The Law Commission's plane example.

\section{Setup and Background}\label{sec:setup}

For the rest of the paper we will discuss the following example. 

A plane owner named Ag has a choice between placing a bomb on their plane or going shopping. If they place the bomb on the plane, the plane will explode midair and they will collect on their insurance policy. The passengers in the plane will die as a consequence of the plane exploding. The plane owner places a bomb on the plane intending to collect the insurance money, but states that they did not desire to kill the passengers. 

Assume that this scenario can be represented with the following binary variables:
\begin{itemize}
    \item $B \in \{1,0\}$ a decision to place a bomb or not
    \item $P \in \{1,0\}$ The bomb being on the plane
    \item $E \in \{1,0\}$ The plane exploding
    \item $I \in \{1,0\}$ The insurance company paying compensation
    \item $D \in \{1,0\}$ All passengers and crew dying.
    \item $S \in \{1,0\}$ The plane owner going shopping or some other non-harmful activity
\end{itemize}

and the following binary random variables:
\begin{itemize}
    \item $U_E \in \{1,0\}$ A Bernoulli distributed random variable with parameter of success equal to the probability of the bomb exploding (only if placed) 
    \item $U_I\in \{1,0\}$ A Bernoulli distributed random variable with parameter of success equal to the probability of the insurance company paying up (only if explosion)
    \item $U_D\in \{1,0\}$ A Bernoulli distributed random variable with parameter of success equal to the probability of people dying (only if the plane explodes)
\end{itemize}

In the manner of \cite{Halpern2016ActualCausality} Structural Equation model (SEM), the variables listed above are related in the following set of structural equations denoted $\mathcal{F}$:

\begin{itemize}
    \item $P=B$
    \item $S=\lnot B$
    \item $E=P\land U_E$
    \item $I=E\land U_I$
    \item $D=E \land U_D$
\end{itemize}

\begin{figure}[!ht]
    \centering
    \includegraphics[width=0.5\textwidth]{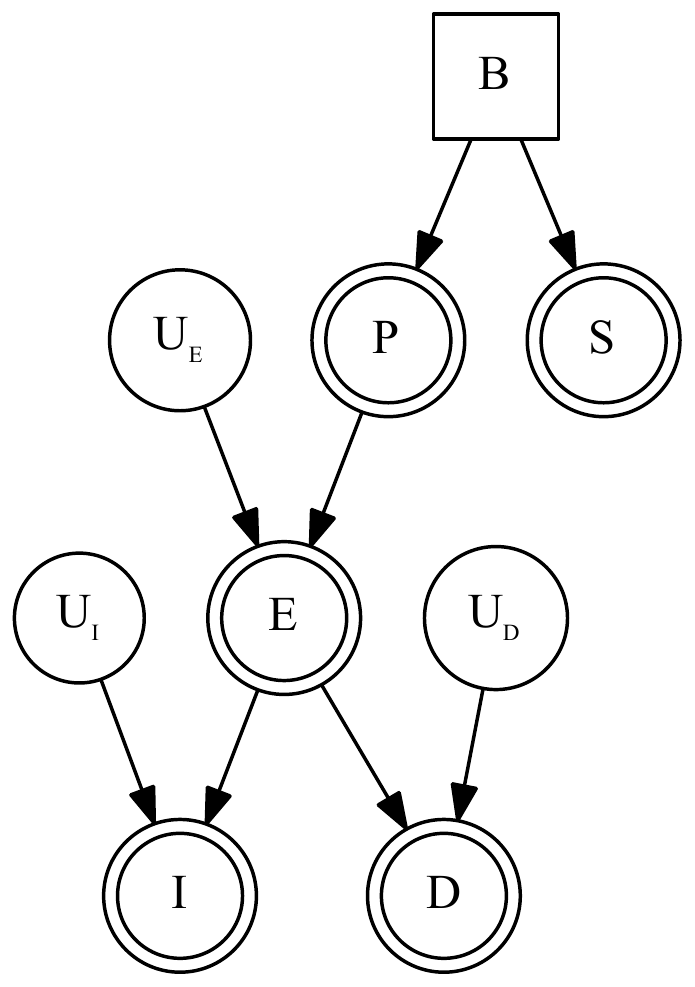}
    \caption{Causal Diagram for the plane bomber. Deterministic variables have double circles. The only decision variable $B$ has a square node. }
    \label{fig:causal_diag}
\end{figure}

The set of variables $\mathcal{U}:=\{U_E,U_I,U_D\}$ are said to be exogenous and the set of variables $\mathcal{V}:=\{B,P,E,I,D,S\}$ are said to be endogenous because they are uniquely defined by the exogenous variables. Action variables are considered endogenous by convention and are determined by an agent. This is to make notation consistent, but one can also assume that once the (possibly non-deterministic) policy is fixed, action variables behave like other endogenous variables. Figure \ref{fig:causal_diag} has one decision variable $B$ which appears exogenous as it has no parents, it could be made to look endogenous by adding an exogenous parental node named $\pi_B$ which would determine a default value for $B$. 

The Signature $\mathcal{S}$ is said to be the triple of $(\mathcal{U},\mathcal{V},\mathcal{R})$ where $\mathcal{R}$ is the set domains for each variable in $\mathcal{U}\cup \mathcal{V}$; we write $\mathcal{R}(U):=Dom(U)$ The causal model $M$ is a signature $\mathcal{S}$ and a set of structural equations $\mathcal{F}$.

More succinctly this can be shown in the causal diagram of Figure \ref{fig:causal_diag}, where a directed edge between variables denotes "is a cause of". Without loss of generality an independent random variable parent can be added for every otherwise non-deterministic variable. This means that the random variables are only ever exogenous and have no parents. It follows that an assignment of variables $\bar{u}$ to the exogenous variables (called a \emph{context} and written $\mathcal{U}\leftarrow \bar{u}$), uniquely determines the endogenous variables $\mathcal{V}$ except the action variables. Conveniently the product distribution of exogenous variables determines the distribution of all variables in the model. A causal model with a context is called a \emph{causal setting}. 

For the causal model $M=(\mathcal{S},\mathcal{V})$, a subset of endogenous variables $X\subseteq\mathcal{V}$, and a particular realisation of those variables $\bar{x}$ the causal model $M_{X\leftarrow \bar{x}}$ is as $M$ with the exception that the equations in $\mathcal{F}$ that determine members of $X$ are replaced by $\bar{x}$. This reflects Pearl's do-algebra \cite{Pearl2000Causality:Inference}; when a variable is intervened on its parental causal relationships are disregarded. In the graph, this equates to its parental arcs being removed. 

Given that all variables are uniquely assigned in a causal setting, then we can define a causal formula $\psi$, a conjunction of assignments or negated assignments of variables in $\mathcal{V}$ to be true or false where $(M,\bar{u})\vDash \psi$ iff $\psi$ is true in the causal setting $(M,\bar{u})$. The relation $\vDash$ is defined inductively, firstly over the variables of the model $(M,\bar{u})\vDash X=x$ iff $X=x$ in causal setting $(M,\bar{u})$, and then in a standard way over negations and conjunctions. We write $(M,\bar{u})\vDash[Y\leftarrow\bar{y}]\psi$ iff $(M_{Y\leftarrow \bar{y}},\bar{u})\vDash \psi$ for any subset $Y\subseteq\mathcal{V}$. In a causal setting $(M,\bar{u})$, where action(s) $a\in \mathcal{R}(A)$ have been chosen we write $(M,\bar{u})_{A\leftarrow \bar{a}}$, 

A utility function in this causal model can be thought of as a function with domain over all variables and a range over the reals; $\textbf{u}:\times_{T\in \mathcal{U}\cup \mathcal{V}}\mathcal{R}(T)\rightarrow \mathbb{R}$, or more simply as a preference ordering over every single possible realisation of $\mathcal{U}\cup \mathcal{V}$. 

The set of all possible causal settings is denoted $\mathcal{K}$, with a probability distribution $\mathbb{P}$. An \emph{Epistemic state} is the tuple $(\mathbb{P},\mathcal{K},\textbf{u})$. It can be thought of a probability distribution over all settings of variables (\emph{worlds}) in the causal model. In this vein, let $w_{M,\bar{u}}$ denote the \emph{world} determined by the casual setting of $(M,\bar{u})$. We let $w_{M,\bar{u},X\leftarrow x}$ denote the unique world where the endogenous variable set $X\subseteq{V}$ have been intervened on and set to $x$. Similarly, in a world with action(s) that have been set $a\in \mathcal{R}(A)$ and endogenous variables $X$ intervened on and set to $x$ we write $w_{M,\bar{u},A\leftarrow \bar{a},X\leftarrow x}$.

\section{A structural causal model derived definition of intent}
\cite{Halpern2018TowardsResponsibility} develop a definition of intent using the language of counterfactuals and structural equation models introduced in section \ref{sec:setup}. Their definition requires the actor to have a utility function over all outcomes and assumes that they are a utility maximiser. We will term this version of intent \emph{HKW intent}.

They separate what it means for an agent to intend an action and an outcome. For an action, they state that it is intended if a) the agent did it, b) the agent could have chosen a different action and c) the action was utility maximising. Since any action might have different, possibly exclusive outcomes, a theory of intent covering endogenous variables $O\subseteq V$, with action $a$ is also needed. The definition depends on a \emph{reference action set} written $ REF(a)\subseteq \mathcal{R}(A)$, the reference set could be all actions, or in the case when the action set is very large, it might be substantially smaller, or even a singleton. Often this reference or default action might be simply the action of not doing anything\footnote{The omission of action can in certain restricted circumstances be sufficient for some crimes typically where the actor has a duty of care towards someone and they are in harms way. Omission also creates problems in the analysis of causality \citep{Lombard2020CausationImpossible}}. 

\begin{definition}\label{defn:affect}\citep{Halpern2018TowardsResponsibility} An agent intends to affect variables $O$ through action $a$ and reference set $REF(a)\subseteq \mathcal{R}(A)$ iff the following conditions are met:
\begin{enumerate}[(a)]
    \item \label{cond:def_effect_1} There exists a superset $O'$ of $O$,  $O'\supset O$ such that 
    \begin{multline}\label{eq:affect}
        \sum_{(\mathcal{M},\bar{u})\in K}\mathbb{P}(M,\bar{u})\textbf{u}(w_{M,\bar{u},A\leftarrow a})\leq \\
        \argmax_{a'\in Ref(a)}\sum_{(\mathcal{M},\bar{u})\in K}\mathbb{P}(M,\bar{u})\textbf{u}(w_{M,\bar{u},A\leftarrow a',O'\leftarrow O'_{A\leftarrow a}})
    \end{multline}
    \item $O'$ is the smallest such superset of $O$ with this property. Equivalently, for all strict subsets of $O^*$ of $O'$, \emph{which also contain $O$}, inequality \ref{eq:affect} is not true.
\end{enumerate}

\end{definition}


This definition look daunting. The weighting over all possible worlds, which necessitates the summation, disappears where the world is deterministic. Consider utility quantity \emph{A} derived from choosing action $a$ in causal setting $(M,\bar{u})$ written $\textbf{u}(w_{M,\bar{u},A\leftarrow a})$.  Now consider utility quantity \emph{B} derived in the same causal setting, where a certain set of variables $O$ retain their values as if $a$ had been taken, but otherwise take the value as if an alternative action $a'$ has been taken, written $\textbf{u}(w_{M,\bar{u},A\leftarrow a',O\leftarrow O_{A\leftarrow a}})$. If utility \emph{B} is bigger than utility \emph{A}, the agent would prefer action $a'$ if it otherwise produced the effect of $a$ within a subset of variables $O$. 

The following two definitions are required for the final definition of intent.
\begin{definition}
We say a causal setting $(M,\bar{u})$ is \emph{possible} in epistemic state $(\mathbb{P},\mathcal{K},\textbf{u})$ if $\mathbb{P}(M,\bar{u})>0$ 
\end{definition}

\begin{definition}
In a causal setting $(M,u)$ we say $o$ is \emph{feasible} under $a$ if $(M,u)\vDash(A\leftarrow a)(O=\hat{o})$ \end{definition}

Now we can present the final definition of intent:

\begin{definition}[HKW Intent]\label{def:intent_halp}\citep{Halpern2018TowardsResponsibility}\\
An agent D intends to bring about $O=\hat{o}$ in $(M,\bar{u})$ given an epistemic state $(\mathbb{P},\mathcal{K},\textbf{u})$ and reference set $REF(a)\in R(A)$ iff the following conditions are met:
\begin{enumerate}[(a)]
\item \textbf{Affect intended variable}: D intended to affect $O$ through action $a$
\item \textbf{$a$ can cause $\hat{o}$}: There exists a \emph{possible} causal setting $(M,\hat{u})$ where $\hat{o}$ is \emph{feasible} under $a$.
\item \textbf{$\hat{o}$ is the best outcome}: For every value $o'$ of $O$ with a possible causal setting $(M',\bar{u})$ which is feasible under $a$ we have: 
$$
\sum_{(M,\bar{u})}\mathbb{P}(M,\bar{u})\textbf{u}(w_{M,O\leftarrow \hat{o},\bar{u}})\geq \sum_{(M,\bar{u})}\mathbb{P}(M,\bar{u})\textbf{u}(w_{M,O\leftarrow o',\bar{u}})
$$
\end{enumerate}
\end{definition}

The definition states that a) intended outcomes can only occur in variables D intends to affect b) D considers $a$ can cause $\hat{o}$ and c) of all of the possible values that $O$ could take as a result of action $a$, $\hat{o}$ would bring about the highest utility. Note that there is no requirement for an intended outcome to actually occur. This is consistent with a legal system that also punishes failed attempts to commit a crime. It is also a desirable property for any system which must plan not to intend to do something.

As the authors admit, the requirement that an intended outcome be the best possible one is quite extreme in some situations; good outcomes might be merely sufficient. More generally, there is a requirement for the utility function to be defined over all possible settings of endogenous variables. This seems problematic in situations that do not make any intuitive sense. Should the agent receive the utility from an insurance payout in the impossible situation where there has been a payout but no explosion? If yes, then does the utility function implicitly encode causal information? Now that we have the full definition of intent we can apply it to the bomb plane example. 
\begin{example}\label{ex:bomb_halp1}
Consider the situation as described in section \ref{sec:setup}. Initially we will only discuss the deterministic world, so the causal setting is $\bar{u}:=(U_E=1,U_I=1,U_D=1)$. The bomb will always explode if the bomb is placed, insurance will pay up on explosion and the passengers will always die on explosion. 

The first question, is by placing the bomb on the plane what variables the agent named Ag attempt to affect according to definition \ref{defn:affect}. The world is deterministic and there are only two actions, so the reference action is not placing the bomb ($B=0$). Assume Ag receives utility of 100 for collecting the insurance, 1 for going shopping, 0 for not going shopping, $k<0$ for the passengers dying and 0 otherwise. Whilst Ag would like to collect insurance and the plane not explode $P(I=1,E=0|U_E=1,U_I=1,U_D=1)=0$. Under  $\bar{u}$ and $B\leftarrow 1$ we will have $P=1,I=1,E=1,D=1,S=0$ with probability 1.

$$\textbf{u}(w_{M,\bar{u},B\leftarrow 1})=100+0+k$$

The set of intended affected outcomes is $\{I,E,P\}$\footnote{It seems to us that an alternative reading of the definition could only include $I$ as an intended variable, it depends on how the utility function is defined, but the subsequent analysis would lead to the same conclusion.} a relevant superset is $\{I,E,P,D\}$ since 

$$\textbf{u}(w_{M,\bar{u},B\leftarrow 0,\{I,E,P,D\}\leftarrow \{I,E,P,D\}_{B\leftarrow 1}})=100+1+k$$

which satisfies Equation \ref{eq:affect}. Condition (b) of Definition \ref{defn:affect} is satisfied since only one variable was added to the intended set, thus it must be minimal.

Part (a) of Definition \ref{def:intent_halp} means that Ag can only possibly intend to affect $\{I,E,P\}$. Part (b) is satisfied as Ag thinks placing the bomb will lead to insurance payment. Part (c) is satisfied because it is the best action under the utility function described. \qed

\end{example}

The result in Example \ref{ex:bomb_halp1} that collecting the insurance is the only intended outcome is coincides with the bomber's assertion that they did not \emph{directly} intend to kill the passengers and crew of the plane. As the criminal justice system has concluded, direct intent is not necessary for serious crimes. Else anyone accused of a crime could claim not to have intended the criminal outcome, but something where it was only a side effect. A concept of oblique intent must also exist, but it needs to be restrained enough to prevent all side-effects of actions being sufficient for criminal liability.  

The following definition extends HKM intent to oblique intent.

\begin{definition}[SCM Oblique Intent]
Suppose an agent intends to bring about $\hat{O}=\hat{o}$ in $(M,\bar{u})$ given an epistemic state $(\mathbb{P},\mathcal{K},\textbf{u})$ and $REF(a)\in\mathcal{R}(A)$ then they \emph{obliquely intend} to bring about side effect $O^*=o^*$ with confidence $0<C<1$ where $O^*\cap \hat{O}=\emptyset$ and $O^* \in \mathcal{V}$  if either condition holds:

\begin{enumerate}[(a)]
    \item \begin{equation}\label{defn:ob}
        \sum_{(M,\bar{u})}\mathbb{P}(M,\bar{u}).\mathbbm{1}[(M,\bar{u})\vDash(A\leftarrow a)(O^*=o^*)]>C
    \end{equation}
    
    \item \begin{equation}\label{defn:ob_obtain}
        \frac{\sum_{(M,\bar{u})}\mathbb{P}(M,\bar{u}).\mathbbm{1}[(M,\bar{u})\vDash(A\leftarrow a)(O^*=o*,\hat{O}=\hat{o})]}{\sum_{(M,\bar{u})}\mathbb{P}(M,\bar{u}).\mathbbm{1}[(M,\bar{u})\vDash(A\leftarrow a)(\hat{O}=\hat{o})]}>C
    \end{equation}

\end{enumerate}

Where $\mathbbm{1}[\psi]$ is the indicator function, taking the value $1$ if logical statement $\psi$ is true, and $0$ if false.

\end{definition}

Unlike the definition of direct intent, oblique intent makes no reference to the utility function of the agent. This might make it easier to apply in practice in cases where oblique intent is sufficient to establish. By enforcing the set of intended variables and obliquely intended variables to be disjoint, obliquely intended results cannot also be directly intended results (or their negation).  

Clause (a) of the definition states that an outcome is obliquely intended, if it occurs with probability over a threshold $C$ in the course of events which might occur if action $a$ were taken with the intention to bring about some other outcome. For Example \ref{ex:bomb_halp1}, this means that $E=1,P=1,S=0,D=0$ are obliquely intended. That the placement of the bomb and the plane exploding are intended should not come as any surprise, since they are necessary events. It seems slightly strange to say that a side effect of placing a bomb on the plane is not going shopping though not particularly important in this example.

Clause (b) of the definition changes the support of the probability distribution that we measure $C$, hence the requirement for a normalising term in the denominator. It considers only the world where the intended outcome occurs, and measures how likely obtaining the obliquely intended outcome is. Consider the following example.

\begin{example}[Unreliable bomb]
Ag places an unreliable bomb with a $1.5\%$ of detonating. In the language of the causal model of Example \ref{ex:bomb_halp1}, $U_E$ is a Bernoulli distributed variable, with chance of success $0.015$. All other variables are as before. In a world where Ag places a bomb, and the bomb explodes, and insurance is collected $I=1$, $E=1$ and $D=1$, thus $D=1$ is obliquely intended according to Inequality \ref{defn:ob_obtain} for any $0<C<1$. Inequality \ref{defn:ob} is not satisfied since the overall chance of Death is small. \qed
\end{example}

\section{An Influence diagram analysis of Intent}
\cite{Kleiman-Weiner2015InferenceMaking} provide a formal definition of intent using counterfactual reasoning and influence diagrams \citet{Koller2009StructuredProblems} which we will call \emph{KGLT intent}. The actor is assumed to have a utility function over all outcomes and for them to be a strict utility maximiser. A motivating setting behind the paper is understanding variations of the trolley problem. Intuitively an agent KGLT-intends an outcome, if their policy counterfactually depends on that outcome being realised.

An influence diagram named $ID$ is a directed acyclic graph with nodes divided between decision nodes $\mathcal{A}$ (rectangles), outcome nodes $\mathcal{X}$ (circles) and utility nodes $\mathcal{U}$ (diamonds). Optionally, outcome nodes which are deterministic (completely defined by their parents' realisations) have a double circle. As with Structural causal diagrams let $\mathcal{R}(Y)$ be the set of possible values that node $Y$ in the $ID$ can take. The three types of nodes form a directed acyclic graph, where a directed arc denotes a causal relationship. Structural equations describe the relation between a child and its parents; unlike with the structural causal model in the previous model, structural equations can include a stochastic element. With no particular loss of generality, we can explicitly split the stochastic elements of any structural equation into separate nodes with no parents to be consistent with structural causal models. 

The following treatment of IDs is adapted from \cite{Koller2009StructuredProblems}. A complete policy (or strategy), denoted $\pi_{ID}$ is a set of functions for every decision node in the ID which map all possible inputs to a decision output. More formally, for a specific decision variable  $D$, a decision rule $\pi_D$ is a conditional probability function which assigns a probability distribution to all possible values $\mathcal{R}(D)$ dependent on the parents of $D$ which we will write $pa(D)$. A deterministic decision rule would apply unit weight to exactly one element in $\mathcal{R}(D)$. A decision rule for every $D\in\mathcal{D}$ is a complete policy.

Let $\xi$ represent a full realisation of all variables in ID, which is called an outcome. Let $\xi_Y$ be the realisation at node $Y$. Then the utility from any outcome is the sum of realisations of all utility nodes: $$U(\xi)=\sum_{U\in \mathcal{U}}\xi_U$$

The probability of any realisation, under a complete policy is: $$ \mathbb{P}(\xi|ID(\pi))=\prod_{Y\in\mathcal{X}\cup\mathcal{U}}\mathbb{P}(\xi_Y|pa(Y),\pi)$$
And the expected utility from any policy is therefore:
$$EU[ID(\pi)]=\sum_\xi \mathbb{P}(\xi|ID_\pi)U(\xi) $$

To aid comparison with the SCM approach in the previous section, we will additionally enforce the following:

\begin{definition}(\citet{Heckerman1994ACausality})\label{defn:howard}
A causal influence diagram ID is in \emph{Howard Canonical Form (HCM)} if every chance node that is a descendant of a decision node is a deterministic node. 
\end{definition}

In practice, conversion to HCM means adding a non-deterministic parent $u_Y$ to any non-deterministic node $Y$ and adjusting the structural equation so that $Y$ is known with certainty if the values of $pa(Y)$ are known. Once the set of non-deterministic nodes $\bar{u}\subset \mathcal{X}$ and policy are determined, all other values in the ID are known. Thus a realisation of the non-deterministic nodes can be thought of a causal setting as in a SCM.

The best foreseen outcome F is defined as the outcome which leads to the highest amount of utility to the agent under a (utility maximising) policy. 
\begin{definition}\citet{Kleiman-Weiner2015InferenceMaking}
The best foreseen outcome F($\pi$), is the outcome of the highest expected utility that is possible following a policy $\pi$.

$$ F(\pi)= \argmax_\xi U(\xi)\mathbb{P}(\xi|ID_\pi)$$

For any node $Y$ in ID, let $Y_F$ denote the value that it takes in F.

\end{definition}

\cite{Kleiman-Weiner2015InferenceMaking} state that foreseen outcomes are intended if the optimal policy under the best foreseen outcome counterfactually depends on them. The definition as presented, is a little ambiguous in the way it talks of node removal, so we present a slightly modified form: 
\begin{definition} [KGLT Intent] Adapted from \citet{Kleiman-Weiner2015InferenceMaking}] 
For an ID, with an optimal policy $\pi^*$, and best foreseen outcome $F(\pi^*)$, an \emph{Intention} I is a subset of nodes and an Intended outcome $o$ is realisation of this set produced by the following procedure:
\begin{enumerate}
    \item Let   $S=\mathcal{U}\cup \mathcal{V}$ and $I=\emptyset$ Remove every node from $S$ that does not have a decision as node as an ancestor.
    For every node remaining node $Y$ in $S$, according to their natural ordering, starting from the last node and proceeding in reverse order:
    \item   Consider the $ID_{Y\neq F}$ defined to be the same as ID except where node $Y$ is restricted to not take the value it does in $F$ regardless of the value of its parents (but its children are governed by the same distributions in ID, albeit renormalised for the absence of $Y_F$).  
    \begin{itemize}
    \item If Y is a chance node: 
    If the optimal policy of $ID_{Y\neq F}$ named $\pi^Y$ is different from $\pi^*$, then add $Y$ to the $I$.
    \item Else Y is a decision node:
    If the optimal policy of $ID_{Y\neq F}$ has a lower expected utility, then add $Y$ to the $I$. 
    \end{itemize}
\end{enumerate}

\end{definition}

\begin{figure}[!ht]
    \centering
    \begin{subfigure}[b]{0.45\textwidth}
        \centering
        \includegraphics[width=\textwidth]{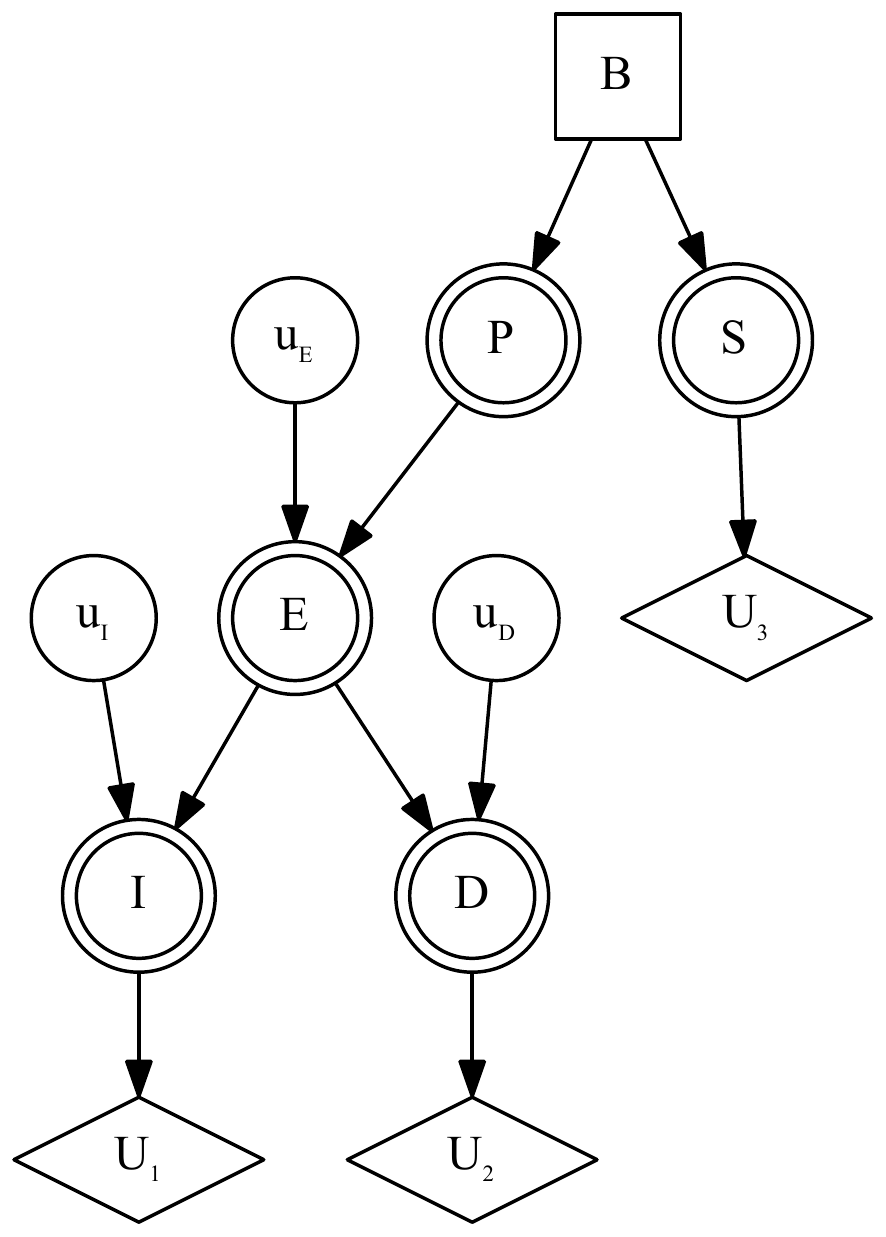}
        \caption{}
        \label{fig:bomb_infl}
    \end{subfigure}
    \hfill
    \begin{subfigure}[b]{0.45\textwidth}
        \centering
        \includegraphics[width=\textwidth]{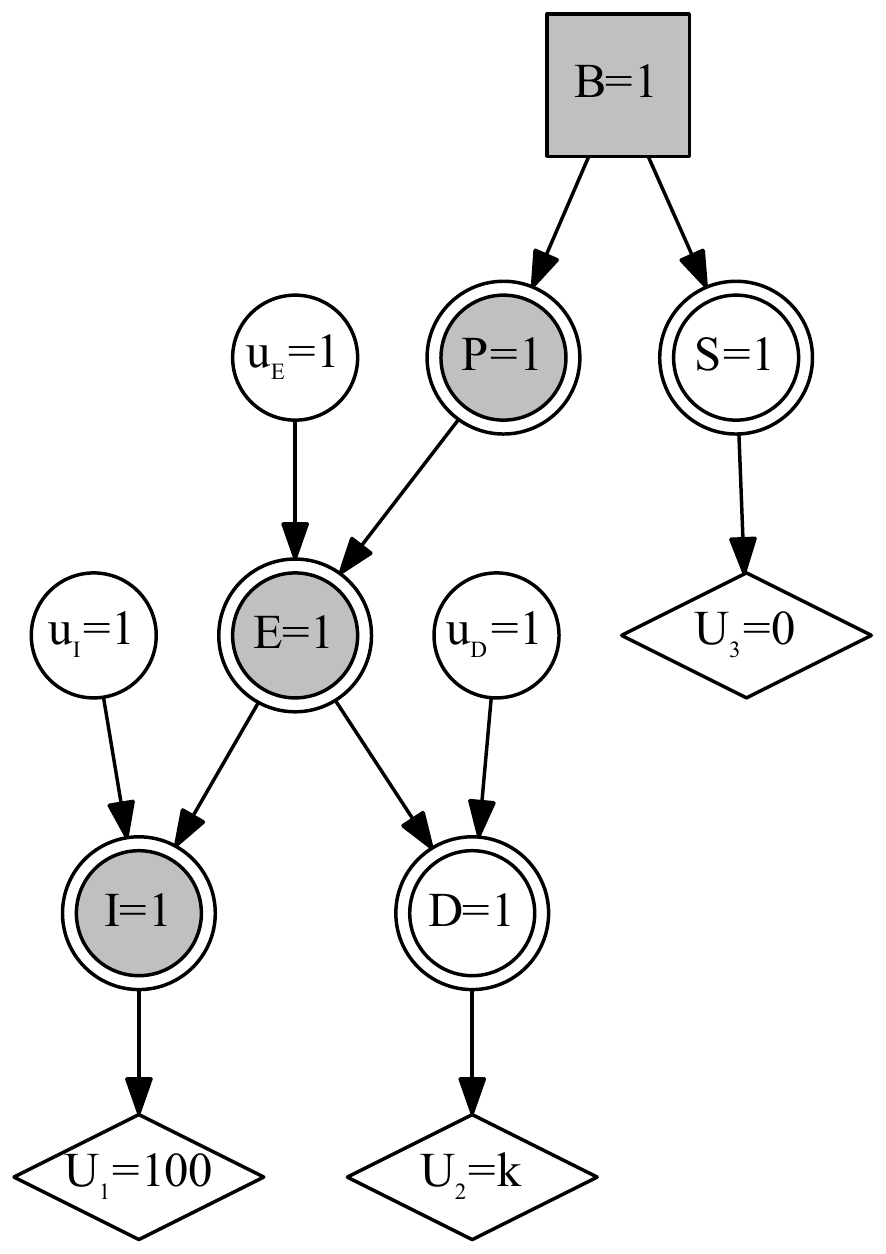}
        \caption{}
        \label{fig:bomb_best}
    \end{subfigure}
    \hfill

    \caption{ Intent in the plane-insurance example using KGLT Intent. (a) Influence diagram representation of problem in Howard Canonical form. (b) Foreseen outcome, corresponding to the variables set to their best outcome according to the agent (example assumes $u_D=1$ is fixed regardless). Intended outcomes are highlighted in grey. Passenger death is not intended. }
    \label{fig:id_bomb}
\end{figure}

\begin{example}\label{ex:bomb_kleim1}
Consider the situation as described in Example \ref{ex:bomb_halp1}. Initially we will only discuss the deterministic world, so the causal setting is $\bar{u}:=(u_E=1,u_I=1,u_D=1)$. The bomb will always explode if the bomb is placed, insurance will pay up if the plane explodes and the passengers will always die. 

The ID is depicted in Figure \ref{fig:bomb_infl}, the best foreseen outcome is shown in Figure \ref{fig:bomb_best}. Starting with node $I=1$, if it was set to $0$, then $U_1=0$, the optimal policy is then to go shopping so $\pi^I\neq \pi^*$, thus $I$ is intended. If $D=0$ $\pi^D= \pi^*$, so it is not intended. We will define $D=1$ shortly as a side-effect. If $E=0$, then $D,I=0$ and $U_1=0$, so $\pi^E\neq \pi^*$, and $E$ is intended, similarly so for $P$. $B$ is a decision node, $EU[ID_{B\neq 1}(\pi^B)]<EU[ID(\pi^*)]$, thus $B=1$ is intended.\qed

\end{example}

We can now extend this account of intent in influence diagrams to oblique intent.

\begin{definition}[ID Oblique Intent]
An outcome $y$ for a variable $Y$ in an Influence diagram $ID$ is obliquely intended with confidence $0<C<1$, with policy $\pi$ and some separate intended outcome $z$ for variable $Z$ if either:
\begin{itemize}
\item \begin{equation}\mathbb{P}(Y=y)=
\mathbb{P}(Y=y|pa(Y),\pi).\prod_{Z\in Anc(Y)}\mathbb{P}(Z|pa(Z),\pi)>C  
\end{equation}
\item Or for some intended result z=Z :
\begin{equation}
    \mathbb{P}(Y=y|Z=z)>C
\end{equation}

\end{itemize}

\end{definition}

As before this definition states that an obliquely intended outcome is one which occurs almost certainly as a result of a policy or almost certainly as a result of an intended outcome being obtained. Once again, no reference is made to the utility function of the agent 

\section{Discussion}
Both HKM and KGLT Intent require the actor to have a known utility function and for them to be utility maximisers. This is a strong assumption as actors may have limited rationality and it might not always be possible to know what their utility function is. HKM Intent makes stronger requirements about the knowledge of the utility function as it requires values to be known for all (possibly nonsensical) scenarios. The influence diagram setting of KGLT intent deals with utility more naturally with the presence of utility nodes. In Example \ref{ex:bomb_halp1} we find that Insurance pay out was intended, and the action of placing the bomb was intended. Whether $E=1$ is intended rather depends on the form of the utility function, for example if $\textbf{u}(I=1,E=1,P=1)=100$ and $\textbf{u}(I=1,E=0,P=0)=0$ then the plane exploding is intended, but wouldn't Ag prefer the insurance payout without the insurance payout?

Unfortunately, oblique intent in Legal literature is often not consistent in its definition. Here we take it to mean the intentional state of non-essential side-effects of actions. Some accounts take it to include necessary intermediate states, caused in realising an intended thing which may not have any particular desirability to the actor, ie the explosion of the bomb in this paper's examples. \cite{Bratman2009IntentionSelf-Governance} terms this means-end intent and \cite{Simester2019MensRea} equates this with direct intent. The effect of this would be to enforce that there exists a path of intended nodes between an action and an intended node in a causal diagram. This seems to be the case of the examples of KGLT intent in \cite{Kleiman-Weiner2015InferenceMaking} though this property of intent is not explicitly listed in desiderata of intent. Perhaps this requirement made explicit might make the analysis of intent quicker in multi-event settings. 

Formal accounts of intent as with those of causality are vulnerable to counter-examples. Far more work exists on the typical types of problems that causal reasoning must overcome (over-determination, preemption, omission) \citet{Liepina2020ArguingArguments}, more work must is needed to mine critical legal discourse on intent and convert it into more formal examples. \cite{Halpern2018TowardsResponsibility} takes great efforts to adapt a theory of intent to account for many possible problems. For example, the additivity of utility can complicate analysis in conjunction with counterfactual reasoning. Adapting an example from Halpern and Kleiman-Weiner, consider the plane example where Ag has two insurance policies. They would like either of the policies to pay out, preferably both. Placing the bomb is not dependent on any single one paying out, since the other may pay. Hence we find that that neither outcome is intended. One approach is to make the definition of an intended outcome more generalised as in Definition \ref{defn:affect}. An alternative might be to transform or restrict the modelling of the problem; events with identical parental causes and utilities of the same sign can be grouped together. It may be that a transformation to a canonical form like that of Howard Canonical (definition \ref{defn:howard}) is needed to make formal intent definitions more robust.  

When considering intent and oblique intent in the wider context of responsibility and culpability, it can become difficult to disentangle the critical elements of each. Some situations may be beyond any theory of intent. The doctrine of Double Effect \cite{McIntyre2019DoctrineEffect} is needed to excuse certain actions and protect, most commonly, physicians from criminal sanction. A surgeon who must perform a risky lifesaving operation will often cause great harm to the patient regardless of whether it is successful and they recover or not. This harm could be viewed as directly intended if invoking means-end intent equivalence to direct intent, or it could be viewed as being obliquely intended. In either case, the surgeon is not culpable; such issues are explored in \cite{Smith1990A}.
The Defence of Necessity\footnote{Ex-US, with the exception of self-defence, which is a puzzle - see \cite{Cotton2015TheLaw}} is another mechanism which exists outside a theory of intent, which excuses intentional actions \citep{Lindsey2011ALaw}. 


\section{Conclusion}
A formal definition of intent is desirable when considering the control and judgement of an algorithmic actor's behaviour. Intent narrowly confined, can label unacceptable outcomes as unintended side-effects. Criminal law acknowledges this in its concepts of Oblique Intent and Knowledge but it is rarer in formal accounts of intent derived from psychology. This paper proposes a definition of oblique intent in two existing formal accounts of intent, which based on counterfactual analysis and the assumption of utility maximising behaviour. The definitions of oblique intent do not require knowledge of the actor's utility function.

\bibliographystyle{spbasic}      

\bibliography{references_excerpt}

\end{document}